\documentclass{article}

\usepackage{arxiv}

\usepackage[utf8]{inputenc} % allow utf-8 input
\usepackage[T1]{fontenc}    % use 8-bit T1 fonts
\usepackage[hidelinks]{hyperref}       % hyperlinks
\usepackage{url}            % simple URL typesetting
\usepackage{booktabs}       % professional-quality tables
\usepackage{amsmath}
\usepackage{amsfonts}       % blackboard math symbols
\usepackage{nicefrac}       % compact symbols for 1/2, etc.
\usepackage{microtype}      % microtypography
\usepackage{lipsum}		% Can be removed after putting your text content
\usepackage{graphicx}
\usepackage{cite}
\usepackage{doi}

\title{Say the Mission, Execute the Swarm: Agent-Enhanced LLM Reasoning in the Web-of-Drones\thanks{This paper has been accepted for presentation at the 27th IEEE International Symposium on a World of Wireless, Mobile and Multimedia Networks (WoWMoM 2026)}}

\date{}

\author{ \href{https://orcid.org/0009-0009-1040-2138}{\includegraphics[scale=0.06]{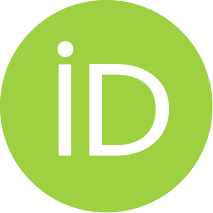}\hspace{1mm}Andrea Iannoli}\\
	Department of Computer Science and Engineering\\
	University of Bologna\\
	Bologna, Italy \\
	\texttt{andrea.iannoli@unibo.it} \\
	\And
	\href{https://orcid.org/0000-0001-9714-3777}{\includegraphics[scale=0.06]{orcid.pdf}\hspace{1mm}Lorenzo Gigli}\\
	Department of Computer Science and Engineering\\
	University of Bologna\\
	Bologna, Italy \\
	\texttt{lorenzo.gigli@unibo.it} \\
	\And
	\href{https://orcid.org/0000-0002-8973-4486}{\includegraphics[scale=0.06]{orcid.pdf}\hspace{1mm}Luca Sciullo}\\
	Department of Computer Science and Engineering\\
	University of Bologna\\
	Bologna, Italy \\
	\texttt{luca.sciullo@unibo.it} \\
	\And
	\href{https://orcid.org/0000-0002-0552-2444}{\includegraphics[scale=0.06]{orcid.pdf}\hspace{1mm}Angelo Trotta}\\
	Department of Computer Science and Engineering\\
	University of Bologna\\
	Bologna, Italy \\
	\texttt{a.trotta@unibo.it} \\
	\And
	\href{https://orcid.org/0000-0001-7496-7597}{\includegraphics[scale=0.06]{orcid.pdf}\hspace{1mm}Marco Di Felice}\\
	Department of Computer Science and Engineering\\
	University of Bologna\\
	Bologna, Italy \\
	\texttt{marco.difelice3@unibo.it}
}

\renewcommand{\shorttitle}{Say the Mission, Execute the Swarm}

\hypersetup{
pdftitle={Say the Mission, Execute the Swarm: Agent-Enhanced LLM Reasoning in the Web-of-Drones},
pdfauthor={Andrea Iannoli, Lorenzo Gigli, Luca Sciullo, Angelo Trotta, Marco Di Felice},
pdfkeywords={Large Language Models,
Model Context Protocol,
Web of Things,
UAV Swarm Control,
Agent-Based Reasoning},
}

\begin{document}
\maketitle

\begin{abstract}
Large Language Models (LLMs) are increasingly explored as high-level reasoning engines for cyber–physical systems, yet their application to real-time UAV swarm management remains challenging due to heterogeneous interfaces, limited grounding, and the need for long-running closed-loop execution.
This paper presents a mission-agnostic, agent-enhanced LLM framework for UAV swarm control, where users express mission objectives in natural language and the system autonomously executes them through grounded, real-time interactions.
The proposed architecture combines an LLM-based Agent Core with a Model Context Protocol (MCP) gateway and a Web-of-Drones abstraction based on W3C Web of Things (WoT) standards. By exposing drones, sensors, and services as standardized WoT Things, the framework enables structured tool-based interaction, continuous state observation, and safe actuation without relying on code generation.
We evaluate the framework using ArduPilot-based simulation across four swarm missions and six state-of-the-art LLMs. Results show that, despite strong reasoning abilities, current general-purpose LLMs still struggle to achieve reliable execution—even for simple swarm tasks—when operating without explicit grounding and execution support. Task-specific planning tools and runtime guardrails substantially improve robustness, while token consumption alone is not indicative of execution quality or reliability.
Overall, the study highlights both the potential and the current limitations of LLM-based swarm control, showing that agent-enhanced execution and standardized device abstractions are essential to translate natural-language intent into dependable swarm behavior.
\end{abstract}

% keywords can be removed
\keywords{Large Language Models \and
Model Context Protocol \and
Web of Things \and
UAV Swarm Control \and
Agent-Based Reasoning}

\section{Introduction}

\subsection{Context}
Unmanned Aerial Vehicle (UAV) swarms are increasingly deployed in applications ranging from precision agriculture and environmental monitoring to search-and-rescue operations and infrastructure inspection \cite{Zeeshan25}. Coordinating multiple heterogeneous drones in dynamic and uncertain environments remains challenging, as it requires reasoning over interacting entities, matching platform capabilities to mission requirements, and adapting to evolving conditions in real time \cite{Trotta18}. Traditional swarm management approaches typically rely on tightly coupled, application-specific integrations that require substantial engineering effort and provide limited flexibility when adapting to new platforms or mission profiles. 

In contrast, Large Language Models (LLMs) have recently emerged as powerful reasoning engines capable of interpreting natural-language instructions, decomposing complex tasks, and orchestrating multi-step workflows. Consequently, their application to cyber-physical and robotic systems has attracted growing interest, as they may bridge the gap between high-level human intent and low-level actuation \cite{Vemprala24,Tian25}. However, their effectiveness and practical usability in such contexts remain largely unexplored. In this paper, we address the following research questions: (\textit{i}) how can LLM-based techniques be exploited for real-time UAV swarm management starting from mission objectives expressed in natural language? and (\textit{ii}) how mature are existing LLM-based agents for managing mission-critical drone applications?

\subsection{Research challenges}
LLM-based techniques have been widely investigated for natural-language drone command interfaces (e.g., \cite{Lillian25,Jiao23}). However, when used for real-time robotic control, several challenges emerge. First, the heterogeneity of drone platforms, sensors, and communication protocols leads to fragmented control surfaces, limiting interoperability and complicating LLM access to drone state information without ad hoc prompt frameworks \cite{Wenhao25_2}. Second, standalone LLM prompting struggles with long-running tasks requiring persistent state tracking and closed-loop real-time interactions, often resulting in repeated code-generation cycles \cite{Wenhao25}. Errors in generated drone behaviors may cause mission failure and raise safety concerns (e.g., collision risks). Finally, the non-deterministic nature of LLM execution complicates evaluation, since identical inputs may produce different control outputs and swarm behaviors. As a result, recent work focuses on metrics such as convergence properties and mission success rates across repeated experiments \cite{Kannan2024,Rahman2025}.

\subsection{Contributions}
While not aiming to solve all the challenges above, this paper investigates AI agents for real-time, mission-agnostic UAV swarm management without human-in-the-loop intervention during execution. We consider a scenario in which a user specifies mission objectives in natural language (e.g., cover an area), while the AI system autonomously coordinates UAV actions from take-off to mission completion.

As main contribution of this study, we propose an agentic LLM-based architecture for long-running drone missions, supporting real-time state retrieval, adaptive decision-making, and contextual reasoning. Unlike most existing LLM applications to robotic swarms, which rely on static code generation at mission initialization, our platform introduces two key innovations. First, it uses LLM reasoning and function-calling mechanisms to coordinate UAV actions without any code generation phase. Second, it enables continuous swarm control and adaptation to real-time feedback from UAVs and the environment through a closed-loop reason–execute–monitor cycle. The interaction between the LLM agent and the physical world is enabled through the Model Context Protocol (MCP) and the W3C Web of Things (WoT) standard \cite{sciullo2022survey}. By modeling each drone as a WoT Thing, the framework allows the LLM to discover, query, and actuate swarm resources through a standardized interface, decoupling mission logic from platform-specific APIs and protocols. The architecture is further extended with an execution-controller agent that monitors runtime conditions and triggers guardrail prompts \cite{Qinghua23} to guide reasoning and ensure safe mission execution.

We also provide an extensive experimental evaluation in a hybrid real–simulated environment by interfacing our platform with ArduPilot framework for realistic flight control and physics simulation. We consider three swarm mission classes: area coverage (with and without planning tools), formation control, and smart irrigation with ground-device communication. For each mission, we evaluate six state-of-the-art LLMs supporting function calling and spanning different scales and deployment characteristics: GPT v5.2, DeepSeek v3.2, GLM v4.7, Grok v4.1 Fast, Claude Haiku v4.5, and Qwen 3 8B.

The results provide several insights into the performance and suitability of LLM for agent-based swarm management. For instance, we notice that in the coverage scenario, integrating external knowledge (e.g., positioning algorithms from the literature) generally improves performance, although not consistently across LLM models. Smaller LLMs  optimized for low-latency inference typically show lower task performance than larger models. Finally, no clear correlation is observed between mission success rate and token consumption (i.e., operational cost).
In short, our contributions are:
\begin{itemize}
    \item A mission-agnostic agent-enhanced LLM architecture for UAV swarm management;
    \item A WoT-based interoperability framework enabling real-time bidirectional communication between the AI agent and physical systems (e.g., UAVs);
    \item A comprehensive empirical evaluation comparing six state-of-the-art LLMs across three swarm-management tasks, analyzing success rate, safety behavior, and token efficiency.
\end{itemize}

The remainder of this paper is organized as follows. Section \ref{sec:rw} reviews related work. Section \ref{sec:architecture} presents the system architecture and operational flow. Section \ref{sec:implementation} describes the implementation. Section \ref{sec:eval} presents the experimental setup and results. Section \ref{sec:conclusion}  outlines future work.
\section{Related Works}
\label{sec:rw}

After impacting several networking domains, LLMs are expected to drive a paradigm shift in mission-critical communications and robotic swarm management \cite{Zeeshan25}. The relevance of this topic is reflected by the rapid growth of research in the last two years, despite the novelty of the underlying technologies. A recent survey on LLM–UAV integration is presented in \cite{Tian25}, highlighting benefits and challenges across navigation, perception, and planning tasks.
In the literature, three main applications of LLMs to UAV swarm management can be identified: bandwidth optimization (e.g., \cite{Fei25}), positioning optimization, and task code generation. Regarding positioning optimization, LLMs are typically used to extract positioning requirements from natural-language user inputs, while low-level drone control relies on conventional methods. For example, \cite{Lillian25} explores LLM-based Drone Delivery as a Service (DDaS) platforms for extracting delivery information from free-text requests. In \cite{Jiao23}, LLMs are used to generate UAV swarm choreographies, where users specify high-level tasks via natural language and a motion-planning framework produces collision-free trajectories. Similarly, FlockGPT \cite{Lykov24} demonstrates LLM-based generation of geometric formations for drone swarms. In \cite{Biao25}, the authors propose SwarmChain, a collaborative LLM inference system leveraging distributed computation across resource-constrained UAVs using model-splitting techniques.

The use of LLMs for generating operational robotic code is most closely related to our work. In \cite{Rahman2025}, classical swarm algorithms (Boids and Ant Colony Optimization) are compared with LLM-generated implementations. Results show that classical Boids outperform LLM-based variants, whereas LLM-based ACO outperforms its classical counterpart, highlighting the importance of prompt design. In this direction, \cite{Wenhao25_2} proposes a prompt framework exposing available APIs, domain constraints, and examples to improve reasoning reliability in LLM-driven swarm control. A closed-loop code-generation framework is proposed in \cite{Wenhao25}, where LLM-generated mission code is validated in simulation and iteratively refined using feedback from a secondary AI agent. However, generated code remains static and does not incorporate real-time feedback. This limitation is partially addressed in \cite{Ishika23}, where LLM prompting relies on program-like descriptions of available actions and environmental objects. Nevertheless, this approach does not leverage interoperability standards such as the W3C WoT, whose potential for automated LLM-based mashup generation has been recently demonstrated in \cite{Fady26}. Finally, \cite{Kannan2024} proposes a unified LLM-driven framework performing multiple stages of multi-robot task planning directly from natural-language instructions. Experimental results on complex tasks show consistent performance across different LLMs, while ablation studies highlight the importance of modular planning stages.
\section{System Architecture}
\label{sec:architecture}

\begin{figure*}[h!]
    \centering
    \vspace{5pt}
    \includegraphics[width=0.9\textwidth]{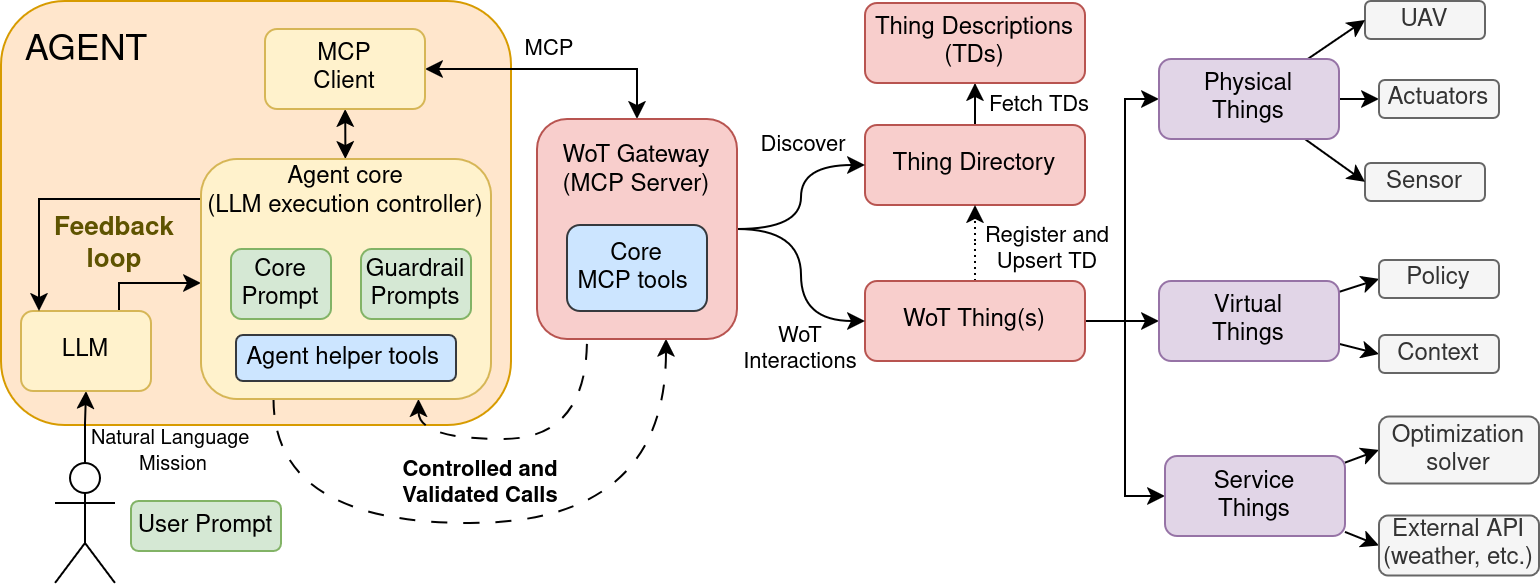}
    \caption{High-level overview of the proposed agent-enhanced, WoT-directed architecture. The Agent encapsulates an LLM and an Agent Core with persistent prompts and guardrails, interacting with a WoT ecosystem through controlled MCP-mediated calls.}
    \label{fig:arch}
    \vspace{-10pt}
\end{figure*}

In this paper, we propose a novel, mission-agnostic software framework that enables the autonomous, LLM-based management of heterogeneous UAV swarms operating in dynamic and uncertain environments. The primary objective of the framework is to allow users to specify high-level mission goals—such as area monitoring, formation control, or distributed data collection—using natural language. The system then autonomously coordinates sensing, actuation, and mobility across multiple aerial devices to execute the mission.  
\newpage
Figure~\ref{fig:arch} provides a high-level overview of the proposed architecture. In contrast to the state-of-the-art approaches from Section~\ref{sec:rw}, our proposal introduces these novelties (\textit{N}):
\begin{itemize}
    \item \textit{N1: Interoperability support}, i.e., the ability to govern heterogeneous drone swarms through a common, hardware-independent interaction model;
    \item \textit{N2: Real-time system state retrieval}, including  swarm drones as well as  external data sources;
    \item \textit{N3: Autonomous closed-loop reasoning}, whereby state information is incorporated into subsequent reasoning steps, enabling multiple phases of tool calling and adaptation during mission execution without human intervention.
\end{itemize}
Items \textit{N1} and \textit{N2} are achieved through the adoption of the W3C Web of Things (WoT) standard \cite{sciullo2022survey} within a Model Context Protocol (MCP) environment \cite{mcp2025spec}. Specifically, the WoT paradigm—standardized by the W3C—provides a uniform abstraction layer through which heterogeneous devices expose their capabilities as Web resources described by machine-readable Thing Descriptions (TDs).  
In the proposed architecture, all system capabilities—including drones, sensors, abstract mission state, and supporting services—are uniformly exposed as WoT Things. %This design provides a common interaction model for swarm-level resources and enables runtime discovery, late binding, and uniform control through standardized interaction patterns. 
As a result, heterogeneous drone platforms and auxiliary services can be seamlessly integrated without requiring modifications to the core swarm-management logic (\textit{N1}).

Moreover, the system state, made accessible through the TDs, is exposed in real time to the LLM via the MCP protocol, an open-source standard that enables AI applications to connect to external data providers. As a result, the swarm can be managed during mission execution through a sequence of  invocations performed by the AI agent (function-calling approach), without requiring any code generation phase (\textit{N2}).

Finally, to enable \textit{N3}, we design an \textit{Agent Core} module that grounds the LLM’s reasoning in the physical world by constraining decision-making to the available WoT affordances. %explicitly exposed by the available WoT resources. 
Concretely, the Agent Core mediates execution through structured tool invocations and responses, thereby limiting the space of admissible interactions and mitigating hallucinated commands,%.% or unsupported assumptions. 
%As a result, once a high-level mission objective is provided, the agent-enhanced LLM autonomously orchestrates discovery, interaction, and adaptation steps through WoT affordances calling, 
while execution feedback is progressively incorporated into subsequent reasoning (\textit{N3}).

In the following, we detail the three main sub-systems of the architecture shown in Figure~\ref{fig:arch}. Finally, in Section~\ref{subsec:flow}, we exemplify the end-to-end execution flow.

\subsection{W3C WoT ecosystem}
The right-hand side of Fig.~\ref{fig:arch} depicts the WoT ecosystem, which provides interfaces to the physical entities (drones, tools, etc). It comprises three main components: Thing Descriptions, the Thing Directory, and the WoT Things.

\textit{Thing Descriptions (TDs)} provide a machine-readable interface specification that describes how a Thing can be discovered and interacted with, including its properties, actions, events (also known as \textit{affordances}), data schemas, and protocol bindings. For instance, individual drone functions (e.g., arming, takeoff, navigation, or mode switching) and sensor operations (e.g., sampling and reporting measurements) can be modeled as actions in the TD.

The \textit{Thing Directory} maintains a registry of available TDs and supports discovery operations, enabling late binding and runtime adaptability.

Finally,  \textit{WoT Things} represent  execution entities exposing their capabilities through the TDs. Each Thing encapsulates the logic required to translate WoT-based affordance interactions into device-specific or service-specific operations while remaining fully decoupled from the LLM-driven reasoning process. In our framework, we support three classes of \textit{WoT Things}: (i) \textit{Physical Things}, such as drones and sensors; (ii) \textit{Virtual Things}, representing abstract or derived entities such as mission state or contextual information; and (iii) \textit{Service Things}, representing computational or informational services such as planners, optimizers, or external APIs. %Swarm-level behavior emerges from the coordinated orchestration of multiple WoT Things by the Agent Core. By composing primitive interactions across multiple entities, the system realizes swarm-level missions while remaining decoupled from platform-specific APIs and communication protocols. At the same time, all devices are accessed uniformly through the same WoT interaction model.

\subsection{MCP-based WoT Gateway}

The MCP-based WoT Gateway acts as the exclusive bridge between the agent-enhanced LLM and the WoT ecosystem. From the Agent’s perspective, it represents the only interface available for discovering and interacting with external resources, thereby enforcing a strictly \emph{WoT-directed} execution model. The MCP-based WoT Gateway provides two core functionalities, exposed as a set of \textit{core MCP tools}.  First, it supports \textit{resource discovery} by querying the Thing Directory to identify available WoT Things that satisfy given requirements. Second, it enables \textit{interaction} by mediating access to WoT Things through the affordances defined in their TDs. Execution results, state updates, and event notifications are propagated back to the Agent via MCP. %This mediation preserves protocol transparency and enforces a clear separation between high-level reasoning and low-level execution.

\subsection{The Agent}

The leftmost component of the architecture in Figure~\ref{fig:arch}, denoted as the \emph{Agent}, represents the central execution entity responsible for interpreting user intents and orchestrating interactions with the external system. The Agent has three components: an LLM, an Agent Core, and an MCP client.

The \textit{LLM} provides high-level reasoning capabilities for interpreting user-defined natural-language missions, such as area coverage, formation control, or coordinated data collection. Rather than generating low-level code for drone commands, the LLM operates at a semantic level, reasoning over abstract concepts such as capabilities, constraints, and execution intent. Its output consists of structured reasoning steps and tool-invocation requests, which are grounded by the Agent Core before interacting with the physical system.

The \textit{Agent Core} acts as the execution controller and contextual runtime for the LLM. It hosts the LLM-driven execution loop and maintains a persistent execution context defined through a \emph{core prompt}. This prompt does not encode control logic or planning algorithms; instead, it establishes semantic boundaries and interaction principles governing the LLM’s behavior, including its role as a drone swarm coordinator and the use of W3C WoT as the interaction paradigm. In addition to the core prompt, the Agent Core incorporates a set of \emph{guardrail prompts} \cite{Qinghua23} that are conditionally activated at runtime. These guardrails are triggered by situations such as incomplete task execution, invalid interaction sequences, or unmet termination conditions. When activated, they are injected into the LLM context to constrain or redirect reasoning, reinforcing execution correctness without hard-coded control policies.

%As a result, the interaction between the LLM and the Agent Core forms a closed \emph{feedback loop}, explicitly depicted in Figure~\ref{fig:arch}. Observations derived from tool invocations and execution feedback are progressively incorporated into subsequent reasoning steps, enabling adaptive behavior during long-running missions.
The \textit{MCP client} provides the communication interface between the Agent and the external execution environment. Finally, the Agent Core integrates a set of \textit{agent helper tools} that extend the basic LLM-driven execution loop with auxiliary computational and control functionalities. These tools address limitations related to reasoning efficiency, numerical processing, and execution semantics that may arise with specific LLM configurations, including edge-oriented models (examples are discussed in Section~V). %Agent helper tools encapsulate reusable, domain-specific functions—such as execution monitoring, completion verification, and controlled synchronization—and invoke the same core MCP tools via the MCP client to ensure validated interactions with WoT Things.

\subsection{End-to-End Operational Flow}
\label{subsec:flow}

\begin{figure}[t]
  \centering\vspace{5pt}
  \includegraphics[width=0.55\columnwidth]{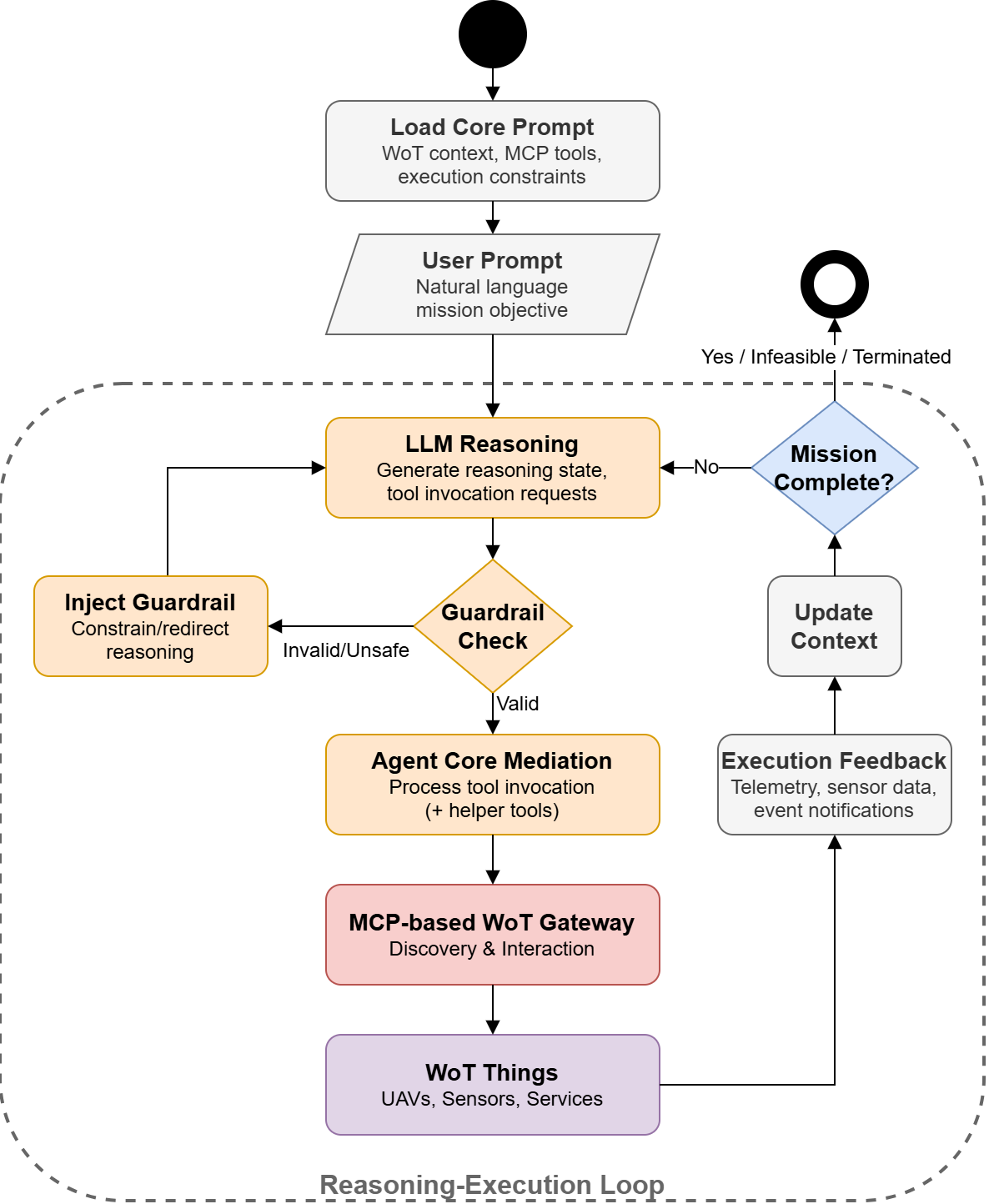}
  \caption{Operational flow of the agent-enhanced swarm mission execution. The iterative reasoning-execution loop integrates LLM-based planning, runtime guardrails, and MCP-mediated WoT interactions, with execution feedback progressively incorporated into subsequent reasoning steps.}
  \label{fig:flow}
  \vspace{-1em}
\end{figure}

From an operational perspective, a swarm mission execution is structured as an iterative reasoning-execution loop managed by the Agent Core (see Figure  \ref{fig:flow}). At initialization, the Agent Core loads the  \emph{core prompt} defining the swarm management context, the WoT-directed interaction model, the available MCP tools, and execution constraints governing LLM behavior. An example of such a constraint is the requirement to guarantee minimum safety distances between drones during mission execution in order to avoid collision events.

A mission is initiated when a user provides a high-level objective expressed in natural language. For instance, a user may define a mission in which all available drones are required to reach a target area to collect data from ground-based IoT sensors. Starting from the \emph{user prompt},   the LLM produces an initial reasoning state that may include inferred objectives, constraints, and intended interactions with the environment. During execution, the LLM issues tool invocation requests (i.e., invocation of WoT Thing affordances) that are mediated by the Agent Core and executed through the MCP-based WoT Gateway. Runtime guardrails are continuously evaluated to constrain or halt interactions in response to invalid requests or unsafe conditions. Moreover, the execution feedback, including telemetry updates, sensor readings, and event notifications, is appended to the interaction context and incorporated into subsequent reasoning steps. Through repeated composition of primitive per-entity interactions, coordinated behavior across multiple drones may emerge over time, enabling the realization of swarm-level missions.
The execution loop continues until the mission objectives are achieved, detected as infeasible under current conditions, or explicitly terminated. 
%This operational model enables flexible and largely automated drone swarm management while maintaining grounded interaction, safety constraints, and a clear separation between semantic reasoning, execution orchestration, and device-level control.

\section{Implementation}
\label{sec:implementation}

This section details the implementation of the architecture introduced in Section~\ref{sec:architecture}.
Following Figure~\ref{fig:arch}, the system is realized as three layers:
(\textit{i}) a \textit{Web-of-Things (WoT) layer} exposing UAVs and auxiliary devices as W3C-compliant Things;
(\textit{ii}) an \textit{MCP-based WoT Gateway} providing typed access to WoT discovery and interaction;
and (\textit{iii}) an \textit{LLM-based Agent} executing the iterative reasoning--action loop detailed in Section \ref{subsec:flow}.
%Each layer is realized as a set of containerized services with well-defined interfaces to ensure modularity, reproducibility, and ease of deployment.\footnote{In compliance with the double-blind policy, code and deployment artifacts will be released upon acceptance.}

\subsection{Web-of-Things Layer}
\label{subsec:implementation-wot}

The WoT layer exposes all physical and virtual resources involved in mission execution as Web Things described through TDs.
Each Web Thing defines its properties, actions, and events, enabling uniform access and runtime discovery.

\subsubsection{WoT UAV Things}

Each UAV is exposed as an individual WoT Thing by a dedicated Servient.
The Servient provides a standardized WoT mobility/telemetry interface and forwards the actual commands through MAVProxy\footnote{https://ardupilot.org/mavproxy/}, which bridges the requests to the ArduPilot simulator via MAVLink. The MAVLink protocol is adopted due to its wide availability, vendor neutrality, and well-defined semantics, enabling interoperability across heterogeneous UAV platforms without exposing low-level flight-control details to the Agent.

The TD of each UAV Thing exposes:
(i) \textit{Actions} for primitive control (e.g., takeoff, goto, land, etc.), and
(ii) \textit{Properties} reporting telemetry and system state (e.g., position, mode, energy, etc.)\footnote{Example of a TD for an UAV: https://doi.org/10.5281/zenodo.18433213}.
Action inputs and state transitions are validated at the Servient boundary to enforce basic execution constraints and prevent ill-defined or unsafe interactions. 
As an example, the \texttt{takeoff} requires a strictly positive \texttt{alt} field, and the servient sets the drone to \texttt{GUIDED} before issuing a takeoff command.
%Such validation ensures that higher-level components always operate on physically meaningful abstractions, independently of the reasoning capabilities of the LLM.

%begin{lstlisting}[caption={Excerpt of a UAV Thing Description}, label=lst:td-drone, language=Json]
%...
%"goto": {
%    description: "GUIDED goto lat/lon/alt (m AGL)",
 %   input: {
  %      properties: {
   %         lat: { type: "number" },
    %        lon: { type: "number" },
     %       alt: { type: "number" },
      %  },
       % required: ["lat", "lon", "alt"],
   % },
%},
%...
%\end{lstlisting}
%In Listing~\ref{lst:td-drone} we show a portion of the TD for the \textit{goto} action.
%Additional sensing and actuation resources are exposed as independent WoT Things and become automatically discoverable once their TDs are registered, without requiring changes to the Agent or the MCP-based WoT Gateway.

\subsubsection{Thing Description Directory}

The Thing Description Directory is implemented using \emph{Zion}~\cite{aguzzi2024zion}, an open-source, W3C-compliant directory supporting scalable TD storage and metadata-based discovery. Zion was selected for its performance and scalability, as demonstrated in prior evaluations against other W3C-compliant directories \cite{aguzzi2024zion}. All Servients register and periodically update their TDs in Zion via standard CRUDL operations.
Discovery queries rely on TD metadata and JSONPath expressions, allowing the Agent to dynamically locate relevant Things without hard-coded identifiers.

\subsection{MCP Tool Gateway}
\label{subsec:implementation-mcp}

The MCP-based WoT Gateway is implemented as a dedicated NestJS server and represents the only integration point between the Agent and the WoT layer.
It exposes a set of \emph{core MCP tools} for discovery and interaction, each defined with explicit input and output schemas and validated at runtime using Zod\footnote{https://zod.dev/}. To support asynchronous and long-running operations, the gateway separates action invocation from execution completion.
Action calls return acknowledgements or TD form references, while completion can be verified by the Agent through subsequent property reads or event notifications.
%The gateway also supports batched and multi-Thing invocations, reducing latency and token consumption in coordinated multi-UAV operations.

\subsection{LLM Mission Agent}
\label{subsec:implementation-agent}

The LLM Mission Agent is developed in Python using the LangChain\footnote{https://www.langchain.com/} framework and provides a concrete implementation of the Agent architecture described in Section~\ref{sec:architecture}. More precisely:

\paragraph{Prompt realization}
The Agent Core maintains three prompt artifacts with distinct lifecycles.
A persistent \emph{core prompt} is loaded at startup and remains fixed throughout a run; it encodes execution invariants identified as critical for reliable operation, including mandatory state verification, safe termination conditions, and exclusive use of MCP-mediated interactions.
A \emph{user prompt} is injected once per run to specify the mission objective.
Both prompts are implemented as static templates and are not modified during execution.
\paragraph{Guardrails derived from execution failures}
Runtime guardrails are implemented as short, targeted prompt fragments that are injected dynamically by the Agent Core.
Unlike the core prompt, guardrails were not designed a priori but derived empirically from repeated failure patterns observed during early deployments and simulation runs.
Specifically, guardrails are triggered when the Agent Core detects conditions such as:
(\textit{i}) inferred task completion without corresponding state verification,
(\textit{ii}) stalled execution characterized by repeated non-progressing tool calls, or
(\textit{iii}) attempted termination while one or more drones remain airborne or armed.
When activated, guardrails explicitly instruct the LLM to re-evaluate system state and complete missing execution steps.
This mechanism improves robustness without embedding task-specific control logic in code.

\paragraph{Helper tools as execution shortcuts}
Agent helper tools are implemented as optional, higher-level wrappers around existing MCP tools.
They encapsulate recurring execution patterns such as synchronization, completion verification, and coordinated multi-drone commands that were observed to cause excessive tool-call verbosity or bookkeeping errors, particularly in smaller or latency-optimized models.
Importantly, the helper tools do not bypass the WoT-directed execution model: internally, they invoke the same validated MCP tools available to the LLM.
%Their availability is explicitly controlled in the evaluation to assess their impact on execution reliability and token efficiency.

%Concrete examples of the prompts, guardrails, and helper tools used in the experiments are reported in Section~\ref{sec:eval}.

\section{Validation and Evaluation}
\label{sec:eval}

We evaluate the proposed framework in a controlled simulation environment and compare the performance of multiple LLMs on representative swarm-management tasks.
The evaluation focuses on metrics related to  correctness, robustness under tool constraints, operational efficiency, and cost, measured as token consumption across different mission configurations.

\subsection{Experimental setup}

All experiments are conducted using a fully containerized deployment of the system described in Sections~\ref{sec:architecture} and~\ref{sec:implementation}, ensuring reproducibility and consistent execution conditions.
UAV behavior is emulated using \emph{ArduPilot Software-In-The-Loop (SITL)}, with one multirotor instance per drone with a unique SYSID, exposing standard MAVLink telemetry, state transitions (e.g., arming, flight modes), and collision detection.

%The simulated fleet runs ArduPilot SITL with one instance per drone SYSID.
MAVProxy multiplexes MAVLink traffic and forwards it to a single \texttt{mavlink2rest} endpoint, providing an HTTP interface for reading vehicle state and issuing commands.

%This architecture keeps the simulation fast to reset and allows multiple WoT Servients to share the same MAVLink transport while addressing different SYSIDs.
%This setup provides realistic flight dynamics and safety constraints without requiring physical hardware.
% MDF: Gia detto nell'implementazione, per me si può togliere
%Each simulated UAV, auxiliary device, and mission planning service is exposed as a WoT Thing via a dedicated Servient, with Thing Descriptions registered in the Zion Thing Description Directory.
%The Agent interacts exclusively through the MCP Gateway, which exposes typed discovery, planning, and interaction tools.
%This enforces a strictly WoT-directed execution model, preventing direct access to device-specific APIs or simulation internals.
Unless otherwise stated, experiments use a swarm of 10 multirotor UAVs operating under clear and stable weather conditions. %Unless otherwise specified, 
Each \textit{(model, experiment)} configuration is repeated for 10 independent runs with identical initial conditions; non-determinism arises exclusively from LLM inference.

%\subsection{Tools available to the Agent}

\begin{table}[ht!]
\centering
\footnotesize
\setlength{\tabcolsep}{4pt}
\renewcommand{\arraystretch}{1.25}
\begin{tabular}{p{0.95\columnwidth}}
\hline
\textbf{Core MCP tools (always enabled)} \\
\hline
\texttt{list\_web\_things}\\
\quad Discover WoT Things registered in the Thing Description Directory. \\

\texttt{read\_web\_thing\_property}\\
\quad Read telemetry and state properties (e.g., position, mode, battery). \\

\texttt{write\_web\_thing\_property}\\
\quad Update configurable properties where supported. \\

\texttt{call\_web\_thing\_action}\\
\quad Invoke WoT actions on UAVs and sensors (e.g., navigation, sensing). \\
\hline
\textbf{Mission planning tools (core tools, experiment-dependent)} \\
\hline
\texttt{plan\_drone\_formation}\\
\quad Compute target positions for geometric formations (line, star, circle). \\

\texttt{plan\_area\_coverage}\\
\quad Generate coverage waypoints and altitude suggestions from region geometry and camera FOV. \\
\hline
\textbf{Agent helper tools (selected models)} \\
\hline
\texttt{send\_drones\_to\_positions}\\
\quad Dispatch multiple navigation commands in a single call. \\

\texttt{wait\_until\_armed / \_arrived / \_landed}\\
\quad Verify execution progress via observed system state. \\
\hline
\end{tabular}
\vspace{3pt}
\caption{Tools exposed to the Agent during evaluation.}
\label{tab:eval-tools}
\vspace{-5pt}
\end{table}

During evaluation, the Agent has access to a predefined set of tools that
% core MCP tools exposed by the MCP-based WoT Gateway. Additionally, helper tools could be implemented in the Agent Core that internally call the core MCP tools.
% Tools 
are grouped into core interaction tools, mission planning tools, and agent helper tools.
The availability of planning and helper tools is explicitly controlled to support ablation studies and to analyze model capabilities. Table~\ref{tab:eval-tools} shows the list of the implemented tools. Finally, due to space constraints, prompt contents are not reported inline\footnote{The complete set of prompts is available here: https://doi.org/10.5281/zenodo.18433213}.

%\subsection{Prompting protocol}

%The Agent operates under a structured prompting protocol composed of three elements:
%(i) a persistent \emph{core prompt}, loaded once per run, defining operating procedures, safety rules, and the WoT/MCP interaction model;
%(ii) a task-specific \emph{user prompt}, specifying the mission objective and constraints; and
%(iii) \emph{runtime guardrails}, injected only when execution failures or premature termination attempts are detected.

%The core prompt is held constant within each experiment family to ensure comparability across models, while user prompts vary according to the task definition.
%Guardrails do not encode task-specific strategies or plans; instead, they restate safety and completion requirements (e.g., collision avoidance, verified landing, and disarming) when execution deviates from expected patterns.

\subsection{Models and experiments}

\begin{table}[ht!]
\centering
\footnotesize
\setlength{\tabcolsep}{6pt}
\renewcommand{\arraystretch}{1.2}
\begin{tabular}{p{0.35\columnwidth}p{0.50\columnwidth}}
\hline
\textbf{Model} & \textbf{Scale and characteristics} \\
\hline
GPT (\textit{GPT~5.2}) & Proprietary frontier-scale model \\
DS (\textit{DeepSeek~V3.2}) & Large proprietary model optimized for tool use \\
GLM (\textit{GLM~4.7}) & Large proprietary bilingual model \\
Grok (\textit{Grok~4.1~Fast}) & Latency-optimized proprietary variant \\
Claude (\textit{Claude~Haiku~4.5}) & Lightweight proprietary model, fast inference \\
QWEN (\textit{Qwen3~8B}) & Open-weight transformer with 8B parameters \\
\hline
\end{tabular}
\vspace{1pt}
\caption{LLMs evaluated in the experiments.}
\label{tab:llm-models}
\vspace{-5pt}
\end{table}

% \todo[inline]{We select models from different vendors and parameter scales, prioritizing robust tool-calling support and diversity in model families and deployment characteristics (e.g., ``fast'' variants vs.\ larger general-purpose models). https://gorilla.cs.berkeley.edu/leaderboard.html}

We evaluate a diverse set of LLMs differing in scale, architecture, and deployment characteristics.
The list of models considered in this paper is reported in Table~\ref{tab:llm-models}\footnote{For proprietary models, vendors do not release detailed architectural specifications or parameter counts; models are thus characterized by their relative scale and intended deployment regime (e.g., frontier-scale vs.\ lightweight, latency-optimized variants).}.
%With the exception of Qwen3~8B, exact parameter counts are not publicly disclosed by vendors and are therefore omitted\footnote{For proprietary models, vendors do not release detailed architectural specifications or parameter counts; models are thus characterized by their relative scale and intended deployment regime (e.g., frontier-scale vs.\ lightweight, latency-optimized variants).}.
The selected models all provide native support for structured function calling/tool invocation, which is a prerequisite for MCP-mediated execution.
In addition, several of the evaluated models consistently appear among the higher-ranked entries in the Berkeley Function Calling Leaderboard (BFCL), which benchmarks LLMs based on their ability to call functions (i.e., tools) in agentic settings\footnote{https://gorilla.cs.berkeley.edu/leaderboard.html, ver. 2025-12-16}.

We consider four different swarm missions, each stressing different aspects of planning, execution, and reasoning:
(\textit{i}) area coverage with a task-specific planning tool;
(\textit{ii}) area coverage without the planning tool (ablation);
(\textit{iii}) formation control with collision-aware motion; and
(\textit{iv}) smart irrigation based on sensor-driven decision making.
The first two experiments form an ablation study on the impact of explicit planning support. We briefly describe each of them.

\subsubsection{Area coverage}

In the area coverage task, the Agent is instructed to cover a rectangular region of area \(A = 400\,\mathrm{m} \times 300\,\mathrm{m}\) using the available UAVs, subject to altitude bounds and camera field-of-view (FOV) constraints.
Coverage is evaluated using a footprint model where each drone \(i\) covers a circular ground area of radius equal to:
$$
r_{\mathrm{fp},i} = h_i \tan(\mathrm{FOV}/2)
$$
with \(h_i\) denoting the flight altitude of drone \(i\).
The reported coverage ratio is computed as:
$$
%\frac{1}{A} \sum_i \pi r_{\mathrm{fp},i}^2.
(\sum_i \pi r_{\mathrm{fp},i}^2) / A
$$
%i.e., overlaps between footprints are not subtracted.

When the \texttt{plan\_area\_coverage} tool is enabled, the Agent can request vantage points that maximize coverage of the target region.
The planner partitions the region into a near-square grid based on the number of available drones and the region aspect ratio, assigning one target (cell center) per drone.
Let \(w_c\) and \(h_c\) denote the width and height of a grid cell in meters.
The half-diagonal of a cell is given by:
$$
%r_{\mathrm{cell}} = \frac{\sqrt{w_c^2 + h_c^2}}{2}
r_{\mathrm{cell}} = \sqrt{w_c^2 + h_c^2}/2
$$
To ensure that a drone positioned at the cell center covers the entire cell, the planner selects an altitude \(a\) such that the camera footprint radius equals \(r_{\mathrm{cell}}\), yielding to:
$$
%a = \frac{r_{\mathrm{cell}}}{\tan(\mathrm{FOV}/2)}.
a = r_{\mathrm{cell}} / \tan(\mathrm{FOV}/2).
$$
The resulting altitude is bounded to mission-defined minimum and maximum limits.
The Agent then executes the plan and completes the mission by safely landing and disarming the swarm. In the ablation condition, the planning tool is disabled and the LLM must derive a coverage strategy directly from the region geometry, altitude bounds, and FOV metadata exposed through WoT properties.

\subsubsection{Formation experiment}

In the formation task, the user prompt requests a star formation using the available drones.
The Agent has access to the \texttt{plan\_drone\_formation} tool, which generates target slot coordinates for multiple geometric formations around a specified center point and orientation.
Given the set of target slots, the  drone-to-slot assignment is solved as a constrained assignment problem over a distance matrix. To explicitly stress collision-avoidance and execution robustness, the assignment strategy is configured to \emph{maximize the total displacement} between drones' initial positions and their assigned formation slots.
As a result, drones are deliberately routed toward distant targets, increasing path length and the likelihood of trajectory intersections.
In addition, a fixed inter-drone spacing of \(5\,\mathrm{m}\) is enforced within the formation, further amplifying the probability of close encounters.
This setup isolates the Agent’s ability to manage coordinated motion and collision avoidance under adversarial yet realistic conditions.

\subsubsection{Smart Irrigation experiment}

The irrigation task involves three ground humidity sensors and one temperature sensor, each exposed as a WoT Thing with position and communication-range metadata.
A drone must be within range before a sensor reading can be retrieved.
After collecting all sensor readings, the Agent must decide whether to trigger an irrigation action from a ground actuator according to the following decision rule: 
$$
\overline{H} \le 57\% \;\lor\; \overline{T} \ge 30^\circ \mathrm{C}
$$
% \begin{equation}
% \overline{H} \le 57\% \;\;\lor\;\; \overline{T} \ge 30^\circ \mathrm{C},
% \end{equation}
where $\overline{H}$ and $\overline{T}$ denote the mean humidity and temperature values, respectively.
More realistic agronomic models could be considered for decision-making in this scenario; however, they are outside the scope of this experiment, which focuses on testing coordinated swarm data collection. Sensor readings are synthetically generated so that the condition is satisfied or violated with comparable probability across runs.
%This experiment evaluates the Agent’s ability to (\textit{i}) collect distributed sensor data, (\textit{ii}) aggregate measurements, and (\textit{iii}) produce and execute a conditional decision based on observed state. %rather than the correctness of the decision rule itself.

\subsection{Evaluation metrics}

\begin{table}[t]
\centering
\footnotesize
\vspace{9pt}
\setlength{\tabcolsep}{6pt}
\renewcommand{\arraystretch}{1.50}
\begin{tabular}{@{}p{0.26\linewidth}p{0.70\linewidth}@{}}
\hline
\textbf{Experiment} & \textbf{Success criteria (pass/fail)} \\
\hline
\raggedright \textbf{Coverage (with tool)} &
All planned coverage slots are reached within distance and altitude tolerances; all participating drones satisfy \texttt{armed=false} and end in \texttt{mode} \(\in\)\{\texttt{LAND}, \texttt{RTL}\} in the final snapshot; zero collisions. \\

\raggedright \textbf{Coverage (no tool)} &
At least one drone both takes off and later lands inside the target rectangle, reflecting the exploratory nature of this ablation condition; zero collisions. \\

\raggedright \textbf{Formation} &
Star formation detected from final drone positions; all participating drones (i.e., drones that took off during the run) satisfy \texttt{armed=false} and end in \texttt{mode} \(\in\)\{\texttt{LAND}, \texttt{RTL}\} in the final snapshot; zero collisions. \\

\raggedright \textbf{Irrigation} &
All required sensor readings are collected (three humidity sensors and one temperature sensor); irrigation is required iff \((\overline{H} \le 57\%) \lor (\overline{T} \ge 30^\circ\mathrm{C})\); success holds iff the irrigation action is triggered exactly when required and all participating drones satisfy \texttt{armed=false} and end in \texttt{mode} \(\in\)\{\texttt{LAND}, \texttt{RTL}\}. \\
\hline
\end{tabular}
\vspace{1pt}
\caption{Binary success criteria used to score runs}
\label{tab:success-criteria}
\vspace{-17pt}
\end{table}

System performance is assessed using the following quantitative metrics:
\begin{itemize}
    \item \textit{Success rate}: fraction of runs satisfying the task-specific completion criteria;
    \item \textit{Execution time} (successful runs only): wall-clock time from mission start to verified completion;
    \item \textit{Energy consumption} (successful runs only): total battery usage in mAh, used as a proxy for resource efficiency;
    \item \textit{Collision count}: number of inter-UAV collisions during the mission execution;
    \item \textit{Token usage}: total LLM token consumption, used as a proxy for reasoning cost (details below).
\end{itemize}

Safety and efficiency objectives—such as avoiding collisions, limiting unnecessary movements, and completing missions promptly—are explicitly stated as operational priorities in the core prompt.
However, quantitative metrics and aggregation formulas are not exposed to the LLM, ensuring that optimization behavior emerges from execution feedback rather than direct access to scoring functions.
The experiment-specific success criteria are summarized in Table~\ref{tab:success-criteria}.

\subsubsection{Token accounting}
%\paragraph{Token accounting}

Token consumption is measured using the usage metadata returned by the LLM API at each agent iteration~\(i\).
An iteration corresponds to a single reasoning--execution cycle, including prompt construction, model inference, and (if produced) tool-call generation.
The total number of tokens consumed in a run is computed as:
\begin{equation}
\label{eq:token-per-run}
T_{\mathrm{run}}
= \sum_{i=1}^{N} \left( T^{(i)}_{\mathrm{prompt}} + T^{(i)}_{\mathrm{completion}} \right)
\end{equation}

The \emph{prompt} term \(T^{(i)}_{\mathrm{prompt}}\) counts all input tokens sent to the model at iteration~\(i\), including the persistent system instructions, the user-defined mission, the accumulated dialogue history, and any tool outputs injected back into the context as observations:
\begin{equation}
\label{eq:token-prompt}
\begin{aligned}
T^{(i)}_{\mathrm{prompt}}
&= T^{(i)}_{\mathrm{system}} + T^{(i)}_{\mathrm{user}} + T^{(i)}_{\mathrm{history}} + T^{(i)}_{\mathrm{toolout}}
\end{aligned}
\end{equation}

The \emph{completion} term \(T^{(i)}_{\mathrm{completion}}\) counts all output tokens generated by the model, including both natural-language reasoning text and structured tool-call payloads (e.g., JSON arguments):
\begin{equation}
\label{eq:token-completion}
\begin{aligned}
T^{(i)}_{\mathrm{completion}}
&= T^{(i)}_{\mathrm{text}} + T^{(i)}_{\mathrm{toolcall}}
\end{aligned}
\end{equation}

Tool execution itself does not directly consume tokens; however, tool outputs are appended to the interaction history and therefore increase the prompt length in subsequent iterations via \(T^{(i)}_{\mathrm{toolout}}\).
This formulation captures the cumulative cost of long-horizon reasoning, repeated state verification, and tool-mediated interaction, enabling fair comparison of token efficiency across models and experimental conditions.

\subsection{Results}
The following section presents results for the three experiments. For clarity and space constraints, figures report only model names (e.g., GPT), while full version details are provided in Table~\ref{tab:llm-models}.

\begin{figure*}[t]
    \centering
    \includegraphics[width=1\textwidth]{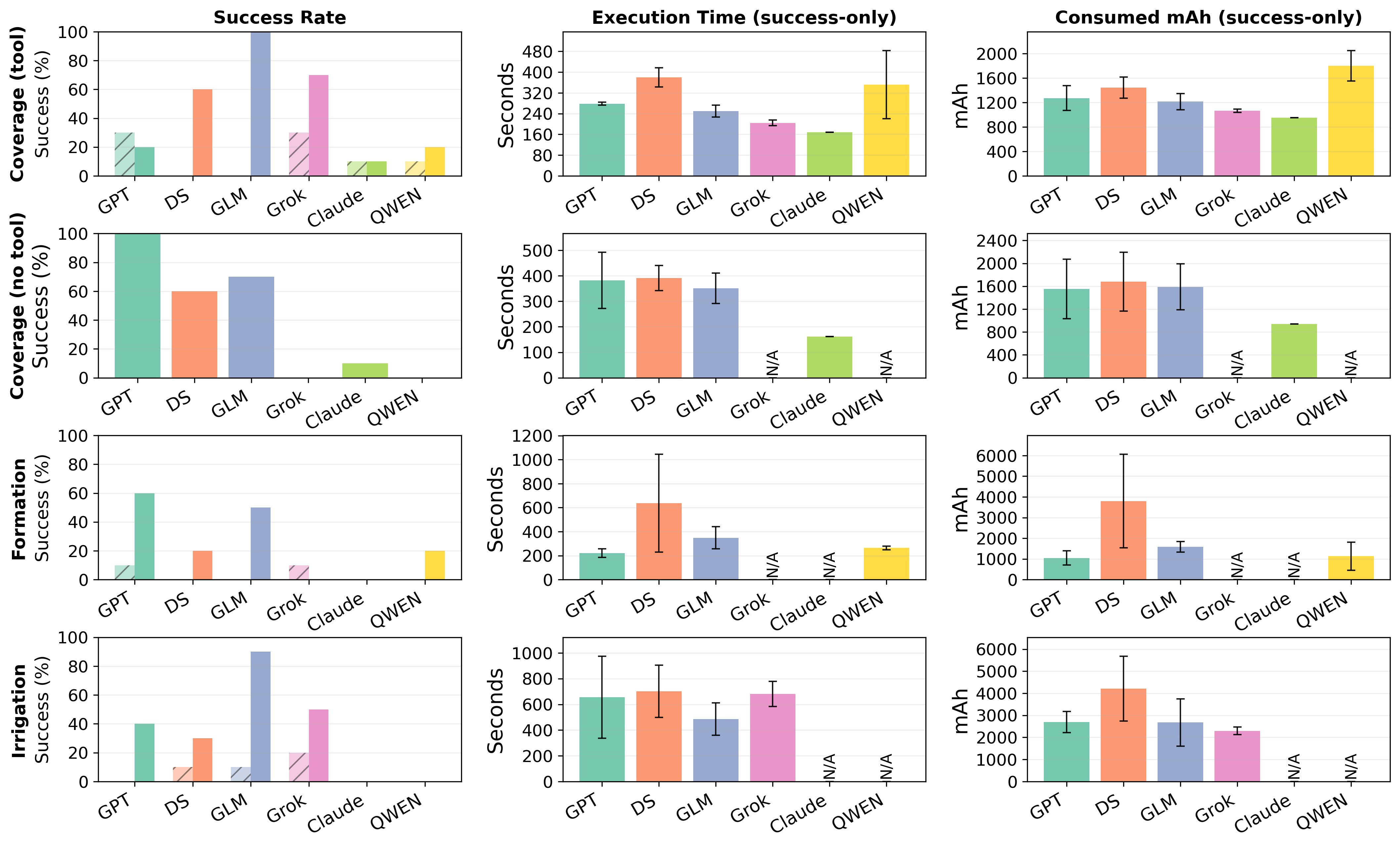}
    \caption{Aggregate results across four swarm-management experiments (rows), comparing models ordered by parameter count (largest$\rightarrow$smallest). \textbf{Left:} success rate, split into \emph{early-exit} successes (hatched bars: task objective achieved but completion constraints not satisfied) and \emph{full} successes (solid bars: task objective achieved and all participating drones end disarmed in \texttt{mode} \(\in\)\{\texttt{LAND}, \texttt{RTL}\}). \textbf{Center/right:} mean end-to-end execution time and mean battery consumption over \emph{full} successful runs only, with error bars denoting $\pm1$ standard deviation; ``N/A'' indicates no successful runs for that model/task. Success is scored using the criteria in Table~\ref{tab:success-criteria} (collisions are counted as failures).}
    \label{fig:summary_bars}
    \vspace{-10pt}
\end{figure*}

\begin{figure*}[t]
    \centering
    \includegraphics[width=1\textwidth]{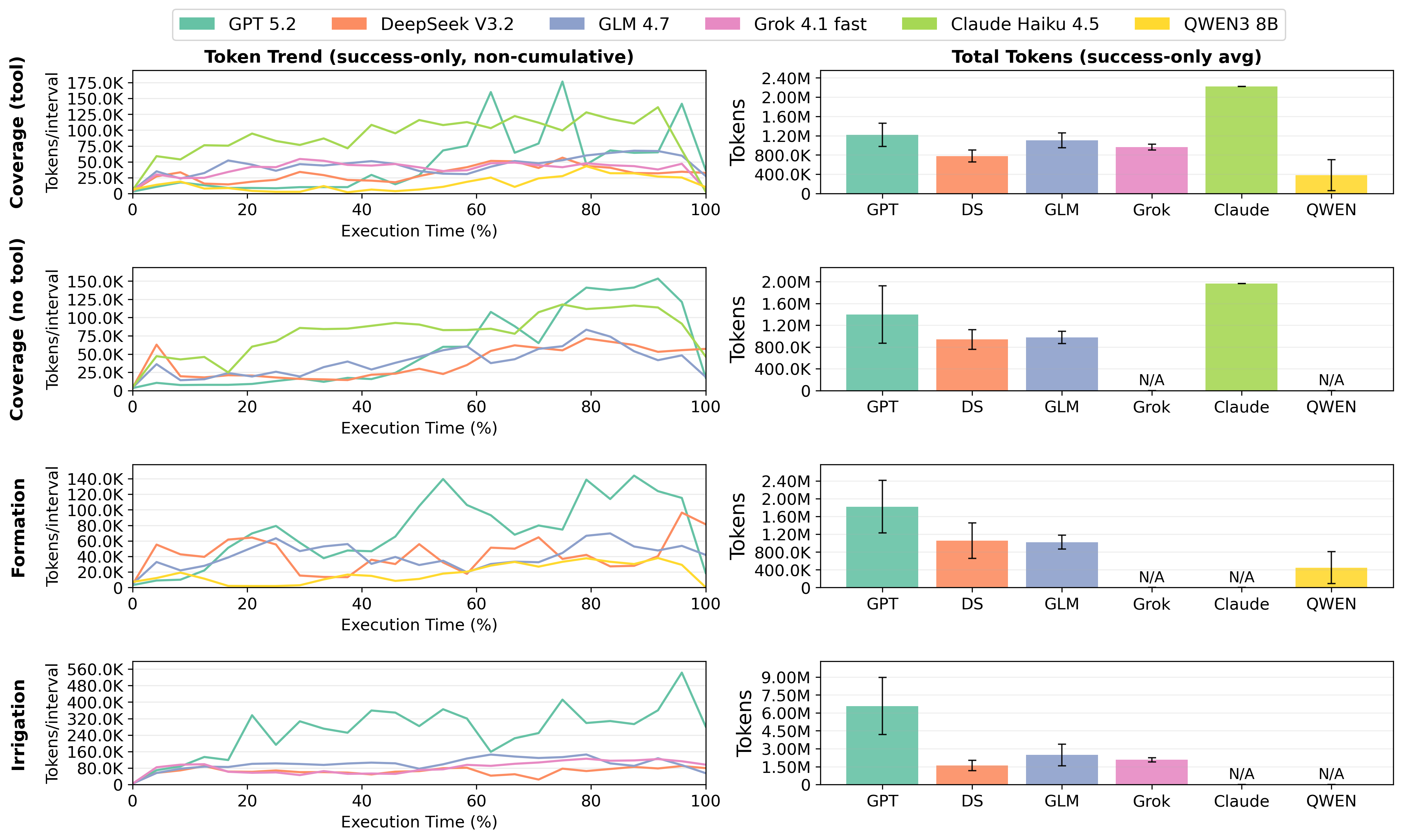}
    \caption{Token-efficiency summary for the same experiments (rows). Left: mean non-cumulative token usage per normalized execution-time interval (0--100\%), obtained by resampling each run's per-iteration token counts to fixed bins and averaging over \emph{full} successful runs. Right: mean total tokens per full successful run with $\pm1$ standard deviation error bars (``N/A'' indicates no successful runs). Token accounting uses provider-reported usage metadata (prompt+completion) as formalized in Eq.~\ref{eq:token-per-run}.}
    \label{fig:summary_tokens}
\end{figure*}

\subsubsection{Area coverage experiment}

Figure~\ref{fig:summary_bars} shows that the task success strongly depends on whether the area-coverage planner tool is available.
In the \emph{with-tool} condition—where the success criterion is stricter and requires all drones to reach their assigned slots \emph{and} to terminate the mission safely (landed and disarmed)—GLM achieves the highest success rate (100\%), followed by Grok (70\%) and DS (60\%).
GPT reaches the planned waypoints in multiple runs but frequently fails to satisfy the final landing and disarming requirements, resulting in a lower success rate (20\%).
In the \emph{no-tool} ablation, which adopts a weaker success criterion (Table~\ref{tab:success-criteria}), GPT achieves a 100\% success rate, DS 60\%, and GLM 70\%.
In contrast, Grok and QWEN do not complete any run successfully (0\%).
For Grok, this behavior is likely related to its reduced reasoning depth as a latency-optimized model, while QWEN appears too small to reliably derive an area-coverage strategy from geometry and FOV constraints alone.
Token usage trends  (Figure \ref{fig:summary_tokens}) are model-dependent.
While the planner introduces additional tool-call and tool-output tokens, it can also reduce the amount of model-generated planning text.
As a result, some models consume fewer tokens when the tool is enabled (e.g., GPT and DS), while others consume more (e.g., GLM and Claude).
A pronounced token spike appears around the first 5\% of execution time in the no-tool condition.
This behavior is consistent with models engaging in an intensive internal planning phase to compensate for the absence of an external planning tool.

Successful executions are generally faster in the with-tool condition (e.g., GLM: $\approx250\,\mathrm{s}$ vs.\ $\approx351\,\mathrm{s}$ without the tool).
Claude represents a notable exception, where the single successful run without the tool completes slightly faster.
This suggests that, for some models, the time spent reasoning about tool usage and handling tool feedback can exceed the cost of computing a coverage strategy internally.
GLM and Claude exhibit decreases in success rate of 30 and 10 percentage points, respectively, when the planner tool is removed.
GPT instead demonstrates an error-prone interaction with the planner, achieving better performance in its absence and reaching a 100\% success rate without the tool.
The smaller model, QWEN, is only marginally capable of completing the mission successfully when a task-specific tool is available and fails entirely when the tool is unavailable.

Overall, the majority of models benefit from explicit planning support, and for several models such tools are essential to mission completion.
GLM demonstrates the most robust behavior across both conditions, achieving a 100\% success rate with the planner and 70\% without it.
Grok attains a 100\% success rate when early exits are counted as successful outcomes, matching GLM when the planner is available; however, Grok is unable to operate without the tool, yielding a 0\% success rate.
\newpage

\subsubsection{Formation experiment}

The formation experiment is more challenging, as it couples multi-drone motion planning with collision avoidance.
GPT achieves the highest success rate (60\%), followed by GLM (50\%).
DS and QWEN complete the mission in 20\% of runs, while Grok and Claude do not complete any run successfully.

\begin{figure}[h]
    \centering
    \includegraphics[width=0.6\textwidth]{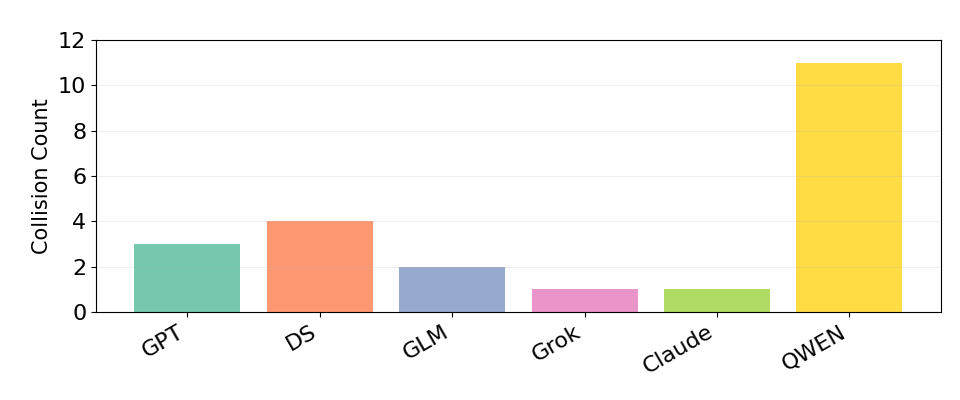}
    \caption{Formation experiment collisions by model. Bars show the total number of collision events detected across all formation runs for each model (lower is better; zero indicates no collisions).}
    \label{fig:formation-collisions}
\end{figure}

Figure~\ref{fig:formation-collisions} reports collision counts aggregated across all formation runs.
A notable comparison emerges between GPT and GLM.
Although GPT achieves a slightly higher success rate, it records three collisions across the ten experimental runs, compared to two collisions observed for GLM.
QWEN exhibits substantially more collisions than all other models, consistent with its reduced ability to manage close-encounter trajectories under the enforced $5\,\mathrm{m}$ spacing and adversarial slot assignment, likely due to its smaller model size.
Grok and Claude rarely collide, indicating that their failures are dominated by other factors, such as incomplete mission termination (e.g., missing landing or disarming) or planning and execution errors, rather than safety violations.

Comparing the two top-performing models, GPT and GLM achieve similar success rates.
However, GLM uses substantially fewer tokens on successful runs (Fig.~\ref{fig:summary_tokens}) and registers fewer collisions across all successful runs.
Overall, while both models perform well, GLM achieves comparable performance with significantly lower reasoning cost and improved safety characteristics.

\subsubsection{Irrigation experiment}

%\begin{figure}[h]
 %   \centering
    %\includegraphics[width=0.48\textwidth]{img/irrigation_success_rate.png}
    %\caption{Aggregate irrigation outcomes by model. Stacked bars report (i) the fraction of runs where all required sensor readings were successfully collected (“data collected only”) and (ii) the fraction of runs where the irrigation decision matched the rule \(\overline{H}\le57\%\ \lor\ \overline{T}\ge30^\circ\mathrm{C}\) (“decision correct”). This figure uses \textbf{partial success} criteria (collection and decision only) and does \textbf{not} enforce end-of-mission completion constraints (e.g., returning/landing/disarming), which are required for full success in the main summaries.}
   % \label{fig:irrigation-success-rate}
%\end{figure}

The irrigation experiment evaluates longer missions requiring the collection of distributed sensor readings under communication-range constraints and the triggering an actuation based on a deterministic decision rule. To be consistent, across runs where all readings are available, the irrigation condition is satisfied in approximately half of the cases, preventing degenerate strategies such as always or never triggering irrigation from achieving high scores.
Figure~\ref{fig:summary_bars} reports \emph{full} success, which requires both a correct irrigation decision and safe mission termination (all participating drones landed and disarmed).
%Figure~\ref{fig:irrigation-success-rate} further decomposes performance into two \emph{partial} milestones: successful data collection and correctness of the irrigation decision, without enforcing end-of-mission constraints.
Under the full-success criterion, GLM achieves the highest success rate (90\%), followed by Grok (50\%).
GPT and DS succeed in 40\% and 30\% of runs, respectively, while Claude and QWEN do not achieve full success.

The partial breakdown explains the observed failures.
GLM successfully collects all sensor readings (100\%) and applies the decision rule correctly in all runs (100\%), with remaining failures attributable to premature termination (decision correct but incomplete landing or disarming).
Grok decides correctly whenever data collection succeeds (70\% decision correctness with 70\% collection), indicating that its dominant failure mode lies in the collection stage (e.g., failure to reach sensor geofences or complete all readings within the run timeout).
GPT and DS both collect all readings in 90\% of runs but achieve only 40\% decision correctness.
Inspection of logged tool calls reveals opposite biases: GPT tends to over-trigger irrigation (false positives), while DS tends to under-trigger (false negatives).
Claude exhibits inconsistent decision behavior even when all data is available.
QWEN fails to reliably reach the data-collection milestone, consistent with its limited ability to sustain long, multi-step tool-based missions.

This task is the most token-expensive in the experimental suite (Fig.~\ref{fig:summary_tokens}), due to repeated navigation, polling, multiple sensing actions, and an explicit decision-making phase.
GPT incurs substantially higher token usage and more agent iterations on successful runs than other models, while GLM again achieves higher success with lower token overhead.
\section{Conclusions}
\label{sec:conclusion}

This paper investigated the feasibility and maturity of agent-enhanced LLMs for real-time UAV swarm management starting from high-level mission objectives expressed in natural language.
We proposed a mission-agnostic architecture in which an LLM performs reasoning and orchestration, while all interactions with physical devices are grounded through standardized W3C WoT affordances and mediated by an MCP gateway.
Unlike code-generation-based approaches, the framework enables continuous closed-loop reasoning and execution via structured tool calls, real-time state observation, and runtime guardrails.

An extensive evaluation across six state-of-the-art LLMs and four representative swarm missions yields several insights.
First, grounding LLM reasoning through typed tools and WoT abstractions is essential for reliable long-running execution, particularly in multi-drone and safety-critical scenarios.
Second, task-specific planning tools significantly improve performance for most models, although their impact varies across architectures.
Third, smaller or latency-optimized models struggle to sustain long-horizon missions without additional execution scaffolding.
Finally, token consumption is not a reliable predictor of mission success, indicating that execution robustness depends more on reasoning structure and feedback integration than on model verbosity.
Overall, the results highlight the need for a dedicated execution layer around the LLM.
Runtime guardrails improve robustness without hard-coding mission logic, while helper tools reduce verbosity and bookkeeping errors for constrained models, enabling safe and transparent operation in cyber--physical systems.

Looking forward, this work suggests a shift in how robotic swarms may be operated.
Rather than programming behaviors or generating mission-specific code, users can increasingly \emph{state intent}, leaving agents to translate it into grounded, verifiable interactions with the physical world.
In this view, drone swarms become executable interfaces to the environment—systems that listen, reason, and act through shared, machine-readable affordances—bringing us closer to a paradigm where saying the mission is enough to make the swarm move.

\bibliographystyle{IEEEtran}
\bibliography{references}  %%% Uncomment this line and comment out the ``thebibliography'' section below to use the external .bib file (using bibtex) .





% \end{thebibliography}

\end{document}